\documentclass[10pt,twocolumn,letterpaper]{article}

\usepackage{cvpr}
\usepackage{times}
\usepackage{epsfig}
\usepackage{graphicx}
\usepackage{amsmath}
\usepackage{amssymb}
\usepackage{bbm}
\usepackage[margin=0pt,font=small,labelfont=small,justification=justified,labelsep=period]{caption}
\usepackage{adjustbox}
\usepackage{multirow}
\usepackage{booktabs}
\usepackage[dvipsnames]{xcolor}
\usepackage{xspace}
\usepackage{color}
\usepackage{multirow}
\usepackage{graphics}
\usepackage{adjustbox}
\usepackage{subfigure}
\usepackage{amsmath}
\usepackage{paralist} %
\usepackage{enumitem} %
\usepackage{booktabs} %
\usepackage{array}
\usepackage{grffile}
\usepackage{makecell}
\usepackage{diagbox}

\graphicspath{{figures/}}

\usepackage[pagebackref=true,breaklinks=true,letterpaper=true,colorlinks,bookmarks=false,citecolor=blue,linkcolor=blue]{hyperref}

\newcommand{\Paragraph}[1]{{\vspace{-2mm}\flushleft\textbf{#1}}} %

\long\def\ignorethis#1{}

\def\OursFullBModuleSmall{inter-pyramid recurrent module}

\def\Ours{BIN}
\def\OursHalfS{BIN}

\def\SpaTime{space-time}

\def\OursConvLstmFirst{$\text{\Ours}_{2}$}
\def\OursConvLstmSecond{$\text{\Ours}_{3}$}
\def\OursConvLstmThird{$\text{\Ours}_{4} \textit{ -w/o cycle loss}$}
\def\OursConvLstmThirdwithCyc{$\text{\Ours}_{4} \textit{ -w/ cycle loss}$}

\def\OursThirdNewLoss{$\text{\Ours}_{4}$}

\newcommand{\leftright}[1]{{#1}}

\def\OursLstmFirst{$\text{\Ours}_{2} \textit{ -LSTM}$}
\def\OursConvLstmFirstY{$\text{\Ours}_{2} \textit{ -ConvLSTM}$}
\def\OursFirst{$\text{\OursHalfS}_{2}$}
\def\OursSecond{$\text{\OursHalfS}_{3}$}
\def\OursThird{$\text{\OursHalfS}_{4}$}

\def\OursFirstNa{$\text{\Ours}_{2} \textit{ -None}$}

\usepackage{tabularx}

\definecolor{B1}{RGB}{237,219,201}
\definecolor{MyBlueb}{RGB}{95,169,236}
\definecolor{MyOrange}{RGB}{255,177,98}
\definecolor{MyReda}{RGB}{193,39,45}
\definecolor{gray}{rgb}{0.5,0.5,0.5}
\definecolor{MyBlue}{rgb}{0,0,1.0}
\definecolor{MyYellow}{rgb}{0.9,0.9,0}
\definecolor{MyRed}{rgb}{0.8,0.2,0}
\definecolor{MyGreen}{rgb}{0,0.5,0.0}
\definecolor{MyGray}{rgb}{0.4,0.4,0.4}

\def\red#1{\textcolor{MyRed}{#1}}
\def\blue#1{\textcolor{MyBlue}{#1}}

\def\first#1{\red{\textbf{#1}}}
\def\second#1{\blue{\underline{#1}}}

\newlength\paramargin
\newlength\figmargin
\newlength\secmargin

\newcolumntype{L}[1]{>{\raggedright\let\newline\\\arraybackslash\hspace{0pt}}m{#1}}
\newcolumntype{C}[1]{>{\centering\let\newline\\\arraybackslash\hspace{0pt}}m{#1}}
\newcolumntype{R}[1]{>{\raggedleft\let\newline\\\arraybackslash\hspace{0pt}}m{#1}}

\def\eg{e.g.,~}

\def\etal{et~al.\xspace}

\newcommand{\secref}[1]{Section~\ref{#1}}
\newcommand{\figref}[1]{Figure~\ref{#1}}
\newcommand{\tabref}[1]{Table~\ref{#1}}
\newcommand{\eqnref}[1]{Equation~\eqref{#1}}

\makeatletter

\setbox0\hbox{$\xdef\scriptratio{\strip@pt\dimexpr
		\numexpr(\sf@size*65536)/\f@size sp}$}

\newcommand{\scriptveryshortarrow}[1][5pt]{{%
		\hbox{\rule[\scriptratio\dimexpr\fontdimen22\textfont2-.2pt\relax]
			{\scriptratio\dimexpr#1\relax}{\scriptratio\dimexpr.4pt\relax}}%
		\mkern-5mu\hbox{\let\f@size\sf@size\usefont{U}{lasy}{m}{n}\symbol{41}}}}

\makeatother

\usepackage[pagebackref=true,breaklinks=true,letterpaper=true,colorlinks,bookmarks=false]{hyperref}

 \cvprfinalcopy %

\def\cvprPaperID{655} %

\ifcvprfinal\fi

\begin{document}

    \title{
    Blurry Video Frame Interpolation\vspace{-3mm}
    }
    
	\author{
	{Wang Shen}
	\hspace{3pt}
	Wenbo Bao
    \hspace{3pt}
	Guangtao Zhai$^{\ast}$
	\hspace{3pt}
	Li Chen
	\hspace{3pt}
	Xiongkuo Min
	\hspace{3pt}
	Zhiyong Gao
	\hspace{3pt}
	\\
	Institute of Image Communication and Network Engineering, Shanghai Jiao Tong University \\
	\vspace{-6mm}
	}
    
    \maketitle
    \thispagestyle{empty}

\begin{abstract}
    Existing works reduce motion blur and up-convert frame rate through two separate ways, including frame deblurring and frame interpolation. 
    However, few studies have approached the joint video enhancement problem, namely synthesizing high-frame-rate clear results from low-frame-rate blurry inputs.
    In this paper, we propose a blurry video frame interpolation method to reduce motion blur and up-convert frame rate simultaneously.
    Specifically, we develop a pyramid module to cyclically synthesize clear intermediate frames.
    The pyramid module features adjustable spatial receptive field and temporal scope, thus contributing to controllable computational complexity and restoration ability.
    Besides, we propose an inter-pyramid recurrent module to connect sequential models to exploit the temporal relationship.
    The pyramid module integrates a recurrent module, thus can iteratively synthesize temporally smooth results without significantly increasing the model size.
    Extensive experimental results demonstrate that our method performs favorably against state-of-the-art methods.
    The source code and pre-trained model are available at
    \url{https://github.com/laomao0/BIN}.{\let\thefootnote\relax\footnote{$^\ast$Corresponding author}}

\end{abstract}

\section{Introduction}\label{intro}

    Shutter speed and exposure time of camera sensors are two fundamental factors that affect the quality of captured videos~\cite{telleen2007synthetic}.
    Slow shutter speed and long exposure time may lead to two kinds of degradations: motion blur and low frame rate.
    Eliminating these degradations is critical for enhancing the quality of captured videos.
    However, few studies have approached the joint problem, namely synthesizing high-frame-rate clear results from low-frame-rate blurry inputs.
    Existing methods may help address this problem by image deblurring and frame interpolation, but are often sub-optimal due to the lack of a joint formulation.

\begin{figure}[t]
	\footnotesize
	\centering
	\renewcommand{\tabcolsep}{0.5pt} %
	\renewcommand{\arraystretch}{0.5} %
	\begin{tabular}{cc}
                \includegraphics[width=0.49\linewidth]{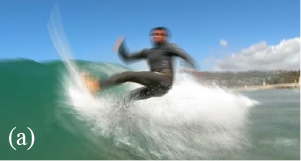} &
                \includegraphics[width=0.49\linewidth]{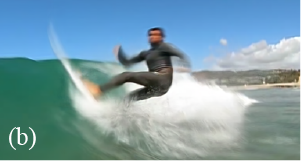} 
				\\
				 \includegraphics[width=0.49\linewidth]{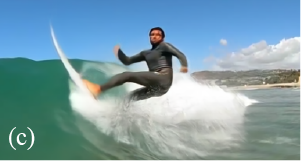} &
				\includegraphics[width=0.49\linewidth]{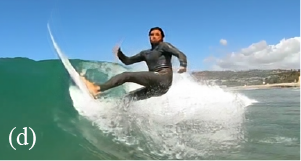} 
				\\
	\end{tabular}
	\vspace{-5pt}
	\caption{
        \textbf{Examples of synthesizing an intermediate frame from blurry inputs.}
        We show the results of 
        (a) overlapped blurry inputs, 
        (b) cascaded interpolation and deblurring model,
        (c) cascaded deblurring and interpolation model, 
        (d) our model.
	}
	\label{fig:preface} %
	\vspace{-10pt}
\end{figure}

    Frame interpolation aims to recover unseen intermediate frames from the captured ones~\cite{bao2019depth,jiang2018super,bao2018memc,bao2018high}.
    It can up-convert frame rate and improve visual smoothness.
    Most state-of-the-art frame interpolation methods~\cite{bao2019depth,jiang2018super,bao2018memc} first estimate objects' motion, and then perform frame warping to synthesize pixels using reference frames.
    However, if the original reference frames are degraded by motion blur, the motion estimation may not be accurate.
    Consequently, it is challenging to restore clear intermediate frames via existing frame interpolation approaches.

    Considering the above problems introduced by motion blur, some existing methods generally employ a pre-deblurring procedure~\cite{tao2018scale, wang2019edvr, su2017deep}. 
    A straightforward approach is to perform frame deblurring,~followed by the frame interpolation, which we refer to as the \emph{cascade} scheme.
    However, this approach is sub-optimal in terms of interpolation quality.
    First, the interpolation performance is highly dependent on the quality of the deblurred images.
    The pixel errors introduced in the deblurring stage will be propagated to the interpolation stage, thus degrading the overall performance.
    Second, most of the frame interpolation methods use two consecutive frames as a reference, namely those methods have a \emph{temporal scope} of two.
    However, given imperfect deblurred frames in the cascade scheme, the interpolation model with a short temporal scope can hardly maintain the long-term motion consistency among adjacent frames.
    An alternative strategy is to perform frame interpolation and then frame deblurring.
    However, the overall quality deteriorates because the interpolated frames suffer from blurry textures of the inputs, as shown in ~\figref{fig:preface}.

    In this paper, we formulate the joint video enhancement problem with a unified degradation model.
    Then we propose a Blurry video frame INterpolation (\Ours) method, including a pyramid module and an inter-pyramid recurrent module.
    The structure of our pyramid module resembles a pyramid that consists of multiple backbone networks.
    The pyramid module is flexible. 
    As the scale increases, the model creates a larger spatial receptive field and a broader temporal scope.
    The flexible structure can also make a trade-off between computational complexity and restoration quality.
    Besides, we adopt cycle loss~\cite{liu2019deep,Reda_2019_ICCV,yuan2018unsupervised,dwibedi2019temporal,wang2018video,pilzer2019refine} to enforce the spatial consistency between the input frames and the re-generated frames of the pyramid module.

    Based on the pyramid structure, we propose an~\OursFullBModuleSmall~which effectively exploits the time information.
    Specifically, the recurrent module adopts ConvLSTM units to propagate the frame information across time.
    The propagated frame information helps the model restore fine details and synthesize temporally consistent images.
    Besides conventional restoration evaluation criteria, we also propose an optical-flow based metric to evaluate the motion smoothness of synthesized video sequences.
    We use both existing databases as well as a new composed dataset crawled from YouTube for performance evaluation.
    Extensive experiments on the Adobe240 dataset~\cite{su2017deep} and our YouTube240 dataset demonstrate that the proposed \Ours~performs favorably against state-of-the-art methods.

    Our main contributions are summarized as follows:
    \begin{compactitem}
        \item
        	We formulate the joint frame deblurring and interpolation problem by exploring the camera's intrinsic properties related to motion blur and frame rate.
        \item 
        	We propose a blurry video frame interpolation method to jointly reduce blur and up-convert frame rate, and we propose an~\OursFullBModuleSmall~to enforce temporal consistency across generated frames.
        \item 
        	We demonstrate that the proposed method can fully exploit~\SpaTime~information and performs favorably against state-of-the-art methods.
    \end{compactitem}

\section{Related Work}
    In this section, we introduce the related literature for frame interpolation, video deblurring, and the joint restoration problem.
    \Paragraph{Video Frame Interpolation.}
        Existing methods for frame interpolation generally utilize optical flow to process motion information~\cite{liu2017video,bao2019depth,bao2018memc,jiang2018super,lee2013frame,xue2019video,niklaus2018context} or use kernel-based models~\cite{long2016learning, niklaus2017video, niklaus2017video2}.
        As a pioneer of learning-based methods, Long~\textit{\etal}~\cite{long2016learning} train a generic convolutional neural network to synthesize the intermediate frame directly.
        The AdaConv~\cite{niklaus2017video} and
        SepConv~\cite{niklaus2017video2} estimate spatially-adaptive interpolation kernels to synthesize pixels from a large neighborhood.
        Meyer~\textit{\etal}~\cite{meyer2018phasenet} use the phase shift of single-pixel to represent motion and construct intermediate frames using a modified per-pixel phase without using optical flow.
        Bao~\textit{\etal}~\cite{bao2018memc} integrate the flow-based and kernel-based approaches.
        Their adaptive warping layer synthesizes a new pixel using a local convolutional kernel where the position of the kernel window is determined by optical flow.
    
        Estimating accurate optical flow is very difficult when the interpolation model encounters blurry inputs.
        We use a variation of the residual dense network~\cite{zhang2018residual} as the \textit{backbone network}. It can generate the intermediate frame without using optical flow.
        Moreover, we use multiple backbone networks to construct a pyramid module, which can simultaneously reduce blur and up-convert frame rate.

    \Paragraph{Video Deblurring.} 
        Existing learning-based deblurring methods reduce motion blur using multiple frames~\cite{su2017deep,jin2018learning,kim2017dynamic,jin2018learning,nah2017cvpr,hyun2015generalized} or single image~\cite{tao2018scale, su2017deep,kupyn2018deblurgan}.
        Wang~\textit{\etal}~\cite{wang2019edvr}~first extract feature information from multiple inputs, then use feature alignment and fusion module to restore high-quality deblurred frames. 
        To further exploit the temporal information, existing algorithms use the recurrent mechanism~\cite{hyun2017online,zhang2018adversarial,sajjadi2018frame,li2019feedback,zamir2017feedback}.
        Kim~\textit{\etal}~\cite{hyun2017online} introduce a spatio-temporal recurrent architecture with a dynamic temporal blending mechanism that enables adaptive information propagation.
        Zhou~\textit{\etal}~\cite{zhou2019spatio} use a spatio-temporal filter adaptive network to integrate feature alignment and deblurring.
        Their model recurrently uses information of the previous frame and current inputs.
        Nah~\textit{\etal}~\cite{nah2019recurrent} adapt the hidden states transferred from past frames to the current frame to exploit information between video frames.
    
        We integrate the backbone network with the proposed~\OursFullBModuleSmall~to operate iteratively.
        The proposed recurrent module adopts ConvLSTM units~\cite{xingjian2015convolutional} to propagate the frame information between adjacent backbone networks.
        Due to the recurrence property, the proposed module can iteratively synthesize temporally smooth results without significantly increasing model size.

    \Paragraph{Joint Video Deblurring and Interpolation.}
        Few studies have approached the joint video enhancement problem.
        Jin~\textit{\etal}~\cite{jin2019learning} introduce the closest related work.
        Their model can be categorized into the jointly optimized cascade scheme.
        It first extracts several clear keyframes, and then synthesizes intermediate frames using those keyframes.
        Their model adopts an approximate recurrent approach by unfolding and distributing the extraction of the frames over multiple processing stages.

        Our method differs from Jin~\textit{\etal}~\cite{jin2019learning}'s algorithm in two aspects.
        First, our model is jointly optimized, and we do not explicitly distinguish the frame deblurring stage or the frame interpolation stage.
        We use the proposed backbone network to associate frame deblurring and interpolation uniformly.
        Second, instead of constructing an approximate recurrent mechanism, we explicitly use the proposed~\OursFullBModuleSmall~that adopts ConvLSTM units to propagate the frame information across time.

\section{Joint Frame Deblurring and Interpolation}
	In this section, we introduce the degradation model for motion blur and low frame rate, and we formulate the joint frame deblurring and interpolation problem.

	\subsection{Degradation Model}\label{model}
\begin{figure}[!t]
	\footnotesize
	\centering	
	\begin{minipage}{1.0\linewidth}
		\centering{ \includegraphics[width= 1.0\textwidth]{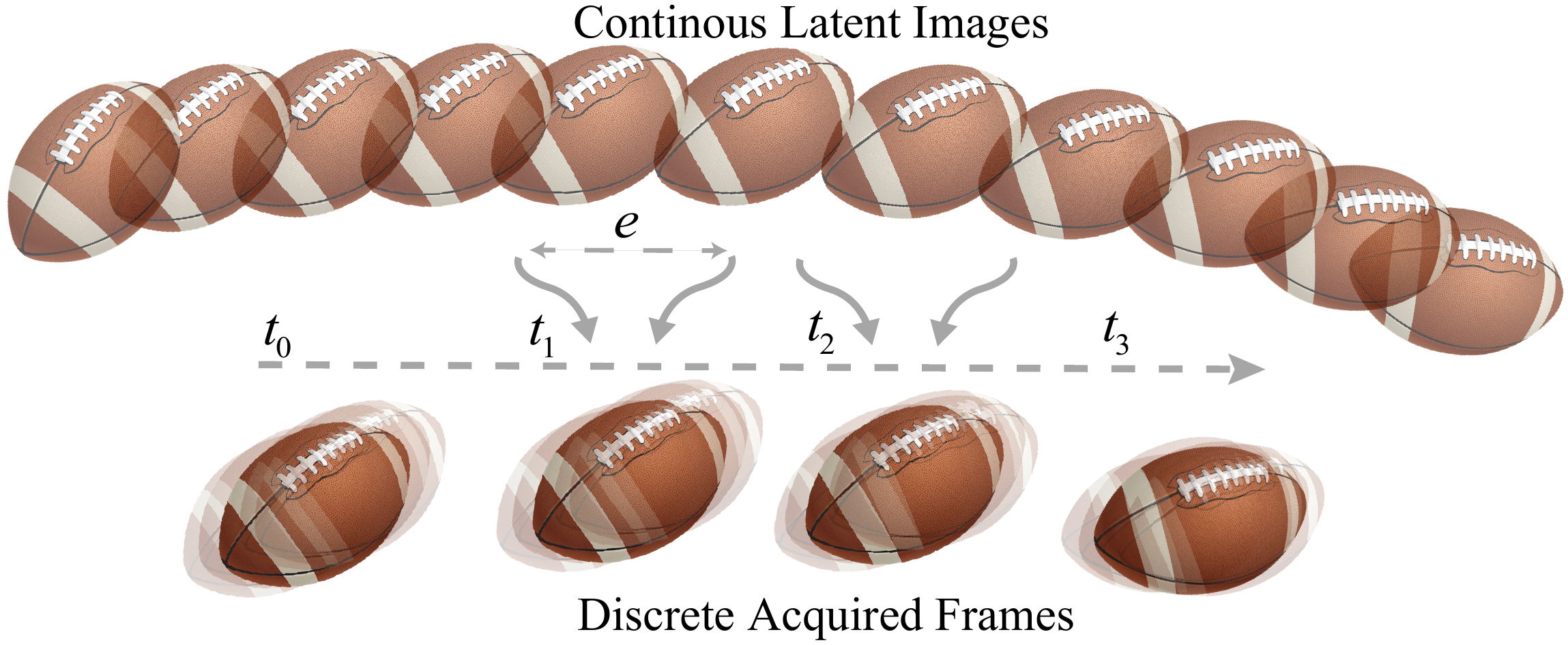}}
	\end{minipage}
	\hfill
	
 	\vspace{-5pt}
	\caption{
        \textbf{Example of frame capturing.}
        Camera sensors capture the discrete frames at time step $t_0$, $t_1$, $t_2$, $t_3$, each of which requires continuous latent images within an exposure time interval of $e$.
	    } 
	\label{fig:ball} 
	\vspace{-15pt}
\end{figure}

        Generally, a camera captures videos by periodically turning on and off its shutter~\cite{telleen2007synthetic}.
        While the shutter is on, also known as exposure, the sensors integrate the luminous intensity reflected by objects to acquire the brightness of objects' pixels.  
        Therefore, the exposure time accounts for the pixel brightness, and the shutter on-off frequency determines the video frame rate.
        Formally, we assume that there exists a latent image $\mathbf{L}(\tau)$ at each instant time $\tau$, as shown in~\figref{fig:ball}.
        We integrate the latent images from time $t_1$ over an interval of time (the exposure interval $e$) to obtain one captured frame.
        We formulate the acquisition of a single frame as:
        \begin{equation}
        \mathbf{B}_{t_1} = \frac{1}{e} \int_{t_1}^{t_1+e} \mathbf{L}(\tau) d \tau. \label{degradation}
        \end{equation}
        Then at the next shutter time $t_2$, the camera generates another frame denoted by $\mathbf{B}_{t_2}$.
        The frame rate of the captured video is defined by:
        \begin{equation}
            f  = \frac{1}{t_2 - t_1}.
        \end{equation}
        Particularly, fast objects movement or camera shake during the exposure time would deteriorate the pixel brightness.
        This deterioration is often in the form of visual blur.

    \subsection{Problem Formulation}
        Given low-frame-rate blurred inputs, we aim to generate high-frame-rate clear outputs. 
        Our goal is to enhance the input video to provide a clear and smooth visual experience.
        We formulate the joint blur reduction and frame rate up-conversion problem as maximizing a \textit{posteriori} of the output frames conditioned on the blurred inputs:
        \begin{equation}
        {\mathcal{F}^{\star}} \doteq
        \underset{\mathcal{F}}{\text{max  }}
        p \big( 
        \mathbf{\hat{I}}_{1:1:2N-1} \big|
         \mathbf{{B}}_{0:2:2N} 
         \big),
        \label{map}
        \end{equation}
        where $\leftright{ \mathbf{B}_{0:2:2N}}$ denotes the low-frame-rate blurry inputs starting from index $0$ to $2N$ with a time step of $2$,~$\leftright {\mathbf{\hat{I}}_{1:1:2N-1}}$ represents the restored and frame rate up-converted results, and $\mathcal{F}^{\star}$ refers to the optimal joint space-time  enhancement model.
        We propose to use trainable neural networks to approximate the optimal model $\mathcal{F}^{\star}$.
        We reformulate the problem in~\eqnref{map} as a minimization of the loss function $\mathcal{L}$ over dataset $\mathcal{S}$:
      	\begin{equation}		
         \begin{split}
        & \underset{\mathcal{F}(\cdot; \Theta)}{\text{minimize\quad}}
         \sum_{s\in \mathcal{S}} \mathcal{L}
         \big(
         \mathbf{\hat{I}}_{1:1:2N-1}
         \big| 
         \mathbf{I}_{1:1:2N-1}
         \big)
        \\
        &\text{subject to\quad}
        \leftright{\mathbf{\hat{I}}_{1:1:2N-1}} =
        \mathcal{F}\big( 
        \mathbf{{B}}_{0:2:2N}
        \big),
        \end{split}
         \label{constraints}
         \end{equation}
        where $\leftright{\mathbf{I}_{1:1:2N-1}}$ denotes the ground-truth frames in the video sample $s \in \mathcal{S}$, and ${\mathcal{F}(\cdot; \Theta)}$ refers to the proposed \Ours~with network parameters $\Theta$.

\begin{figure*}
	\footnotesize
	\centering
	\renewcommand{\tabcolsep}{0.1pt} %
	\renewcommand{\arraystretch}{0.01} %
		\begin{tabular}{cc}
			\includegraphics[width=0.495\linewidth]{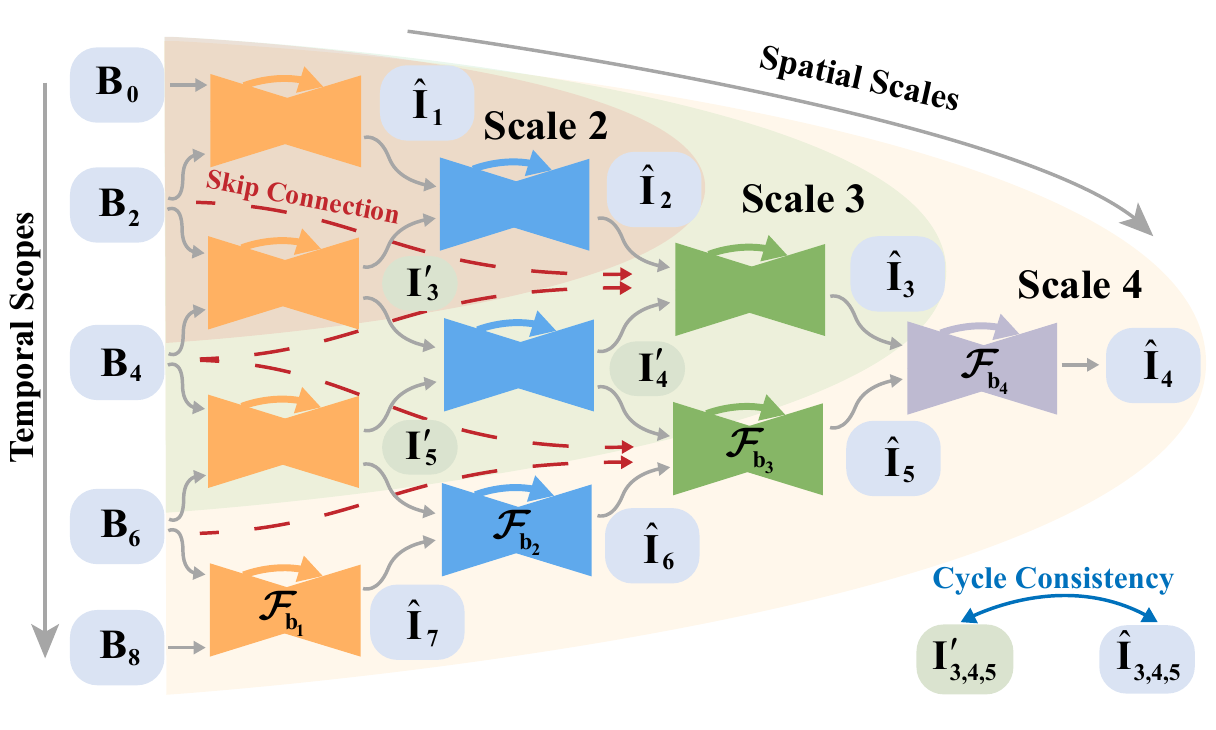}&
			\includegraphics[width=0.495\linewidth]{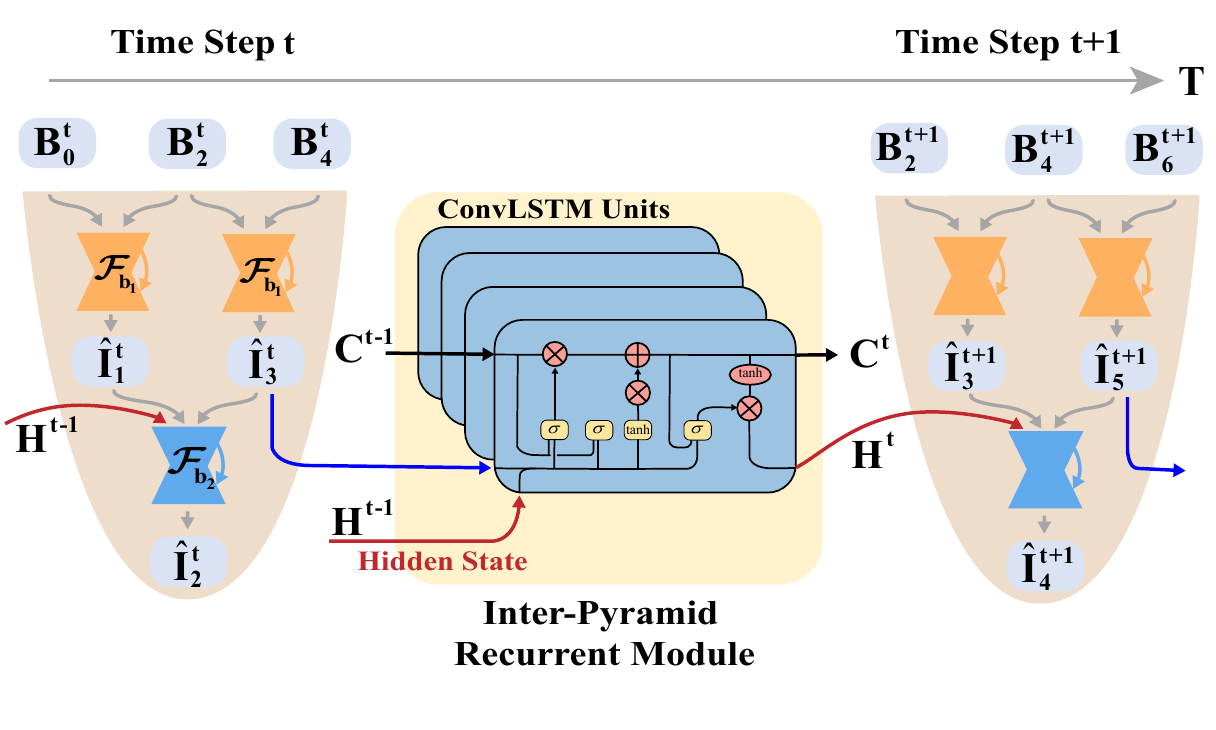}
			\\
			(a)  Pyramid Module & 
			(b)	 Computation Flow of $\text{BIN}_2$ \\ 
		\end{tabular}

	\vspace{-5pt}
    \caption{
        \textbf{Architectures of the proposed blurry video frame interpolation model.}
        The proposed \Ours~consists of two key components, the pyramid module and the~\OursFullBModuleSmall.
        The pyramid module in (a) consists of multiple backbone networks.
        The backbone networks in the same color have shared weights.
        It takes two consecutive frames as input, then synthesizes one intermediate frame.
        The pyramid module can reduce motion blur and interpolate intermediate frames simultaneously.
        Based on the pyramid module in (a), we integrate it with~\OursFullBModuleSmall~to realize the recurrent mechanism in (b).
        The proposed inter-pyramid recurrent module uses ConvLSTM units to propagate the frame information across different pyramid modules.
     }
   
	\label{fig:allstage}
	\vspace{-10pt}
\end{figure*}

\section{Blurry Video Frame Interpolation}

    The proposed model consists of two key components: the pyramid module~and~the~\OursFullBModuleSmall.
    We use the pyramid module to reduce blur and up-convert frame rate simultaneously.
    The~\OursFullBModuleSmall~can further enforce temporal consistency between neighboring frames.
    We show the overall network architecture in~\figref{fig:allstage}.
    Below we describe the design of each sub-network and the implementation details.

\subsection{Pyramid Module}
    The proposed pyramid module integrates frame deblurring and frame interpolation by the following operation:
    \begin{align}\label{eqn:subnetwork}
    \leftright{\mathbf{\hat{I}}_{1:1:2N-1}}
    &= \mathcal{F}
    \big( 
         \leftright{\mathbf{{B}}_{0:2:2N}}
    \big),
    \end{align}
    where $\mathcal{F}$ refers to the pyramid module.
    It takes $\leftright{N+1}$ frames ${\mathbf{{B}}_{0:2:2N}}$ as input, and outputs the deblurred and the interpolated frames $\leftright{\mathbf{\hat{I}}_{1:1:2N-1}}$.
    We construct multiple \emph{backbone networks} to build the pyramid module, as shown in~\figref{fig:allstage}(a).
    The backbone network $\mathcal{F}_{\text{b}}$ interpolates an intermediate frame using two consecutive inputs:
    \begin{equation}
    \mathbf{\hat{I}}_{1} = \mathcal{F}_{\text{b}}(\mathbf{B}_0, \mathbf{B}_2).
    \end{equation}

    The pyramid module has an adjustable spatial receptive field and temporal scope by alternating the scales of the model architecture.
    We show networks with three different scales in~\figref{fig:allstage}(a), denoted by Scale 2,~Scale 3~and~Scale 4.
    The increase of scales makes the entire network deeper, thus creating a larger spatial receptive field. 
    At the same time, the increase of scales also extends the number of inputs, namely the temporal scope, which facilitates the utilization of contextual temporal information.
    For example, the module of scale 2 has a temporal scope of three, while the module of scale 4 can exploit information from five frames, and it has a deeper receptive field compared to the module of scale 2.
    
    Besides the output frames $\leftright{\mathbf{\hat{I}}_{1:1:2N-1}}$, the pyramid module also generates multiple temporary frames.
    As shown in~\figref{fig:allstage}(a), the pyramid module with scale 4 has three temporary frames $\{\mathbf{I}'_{3},\mathbf{I}'_{4},\mathbf{I}'_{5}\}$.
    We use a cycle consistency loss to ensure the spatial consistency between temporary frames with the cycle-paired frames (\eg $\{\mathbf{I}_{3}',\mathbf{\hat{I}}_{3}\}$).

    \subsection{Inter-Pyramid Recurrent Module}
        Temporal motion smoothness is a critical factor in affecting human visual experiences. 
        Based on the pyramid structure, we propose an \OursFullBModuleSmall~to construct multi-scale Blurry frame INterpolation models, denoted by $\text{BIN}_l$, where $l$ is the scale of the pyramid structure.
        The recurrent module can further enforce temporal motion consistency between neighboring frames.
        The~\OursFullBModuleSmall~consists of multiple ConvLSTM units.  
        Each ConvLSTM unit uses hidden states to propagate previous frame information to the current pyramid module.

        For brevity, we illustrate the computation flow of~\OursFirst, which takes one ConvLSTM unit and one pyramid module with scale 2.
        As shown in~\figref{fig:allstage}(b), at time  $t \in [1,{T}]$, given three inputs $\mathbf{B}_{0:2:4}^t$,
        we first generate two intermediate frames $\mathbf{\hat{I}}_{1}^t$ and $\mathbf{\hat{I}}_{3}^t$ by feed-forwarding network $\mathcal{F}_{\text{b}_1}$ twice:
        \begin{align}
            \mathbf{\hat{I}}_{1}^t
              &
            = \mathcal{F}_{\text{b}_1}(\mathbf{B}_{0}^t, \mathbf{B}_{2}^t), \label{eqn:fb1}
            \\
            \mathbf{\hat{I}}_{3}^t 
            &
            = \mathcal{F}_{\text{b}_1}(\mathbf{B}_{2}^t, \mathbf{B}_{4}^t). \label{eqn:fb1_2}
         \end{align}
        Then, we use the synthesized intermediate frames $\mathbf{\hat{I}}_{1}^t, \mathbf{\hat{I}}_{3}^t$ as well as the hidden state $\mathbf{H}^{t-1}$   to synthesize the deblurred frame $\mathbf{\hat{I}}_{2}^t$.
        We extend the backbone network $\mathcal{F}_{\text{b}_2}$ to take the previous hidden state as input, which can be formulated by:
      
        \begin{equation}
            \mathbf{\hat{I}}_{2}^t 
            = \mathcal{F}_{\text{b}_2}( 
            \mathbf{H}^{t-1},\mathbf{\hat{I}}_{1}^{t}, \mathbf{\hat{I}}_{3}^{t}).\label{eqn:fb2}
        \end{equation}
        Besides synthesizing the target frames, the ConvLSTM module also requires to maintain its cell state for temporal recurrence.
        We formulate the updating equation of the inter-pyramid recurrent module by:
        \begin{equation}
        \mathbf{H}^{t}, \mathbf{C}^{t} = \mathcal{F}_{\text{c}}(\mathbf{\hat{I}}_3^t, \mathbf{H}^{t-1},\mathbf{C}^{\leftright{t-1}}) \label{convlstm},
        \end{equation}
        where $\mathcal{F}_{\text{c}}$ refers to the ConvLSTM unit, $\mathbf{C}^{t-1}$ and $\mathbf{C}^{t}$ are previous cell state and current cell state, $\mathbf{H}^{t}$ refers to the current hidden state, and $\mathbf{\hat{I}}_3^t$ denotes the current input.
        At time $t$ and $t+1$, we obtain $\{\mathbf{\hat{I}}_1^{t},\mathbf{\hat{I}}_2^{t}\}$ and $\{\mathbf{\hat{I}}_3^{t},\mathbf{\hat{I}}_4^{t}\}$, respectively.
        By extending the iteration to time $T$, we can synthesize all the deblurred and interpolated frames $ \leftright{\mathbf{\hat{I}}_{1:1:2N}}$.
        
        Following the computation flow of \OursConvLstmFirst, we can extend networks with larger scales~(\eg \OursSecond, \OursThird).
        The network with large scales can utilize a wide receptive field and a broad temporal scope to exploit time information, which can synthesize temporally smooth results.

    \subsection{Implementation Details}\label{imp_details}
    
    \Paragraph{Temporal Skip Connection.}
        We use multiple identity skip connections to pass the pre-stage frame information into later backbone networks, as shown in~\figref{fig:allstage}(a).
        We use identity skip connections to regulate the flow of frame signals for better gradient backward propagation.
        Take~\OursSecond~as an example, the identity skip connections concatenate the inputs $\{\mathbf{B}_{2}, \mathbf{B}_{4}\}$ and the synthesized frames $\{\mathbf{\hat{I}}_{2}, \mathbf{\hat{I}}_{4}\}$ to help the network $\mathcal{F}_{\text{b}_3}$ synthesize the frame $\mathbf{\hat{I}}_3$.

\begin{figure}[t]	 
	\footnotesize
	\centering
    \vspace{-5pt} 
	\begin{tabular}{cc}
		\includegraphics[height=0.95\linewidth]{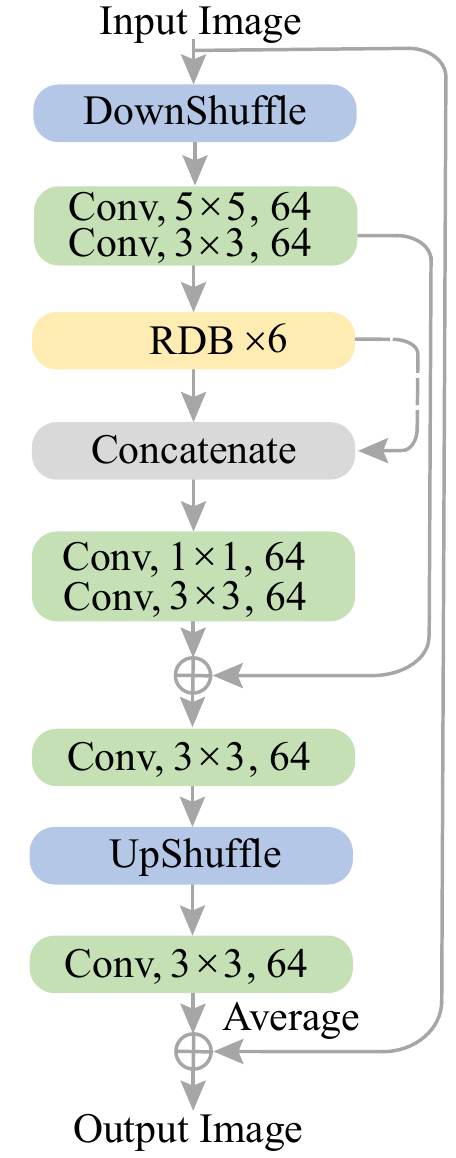}
		&
		\includegraphics[height=0.95\linewidth]{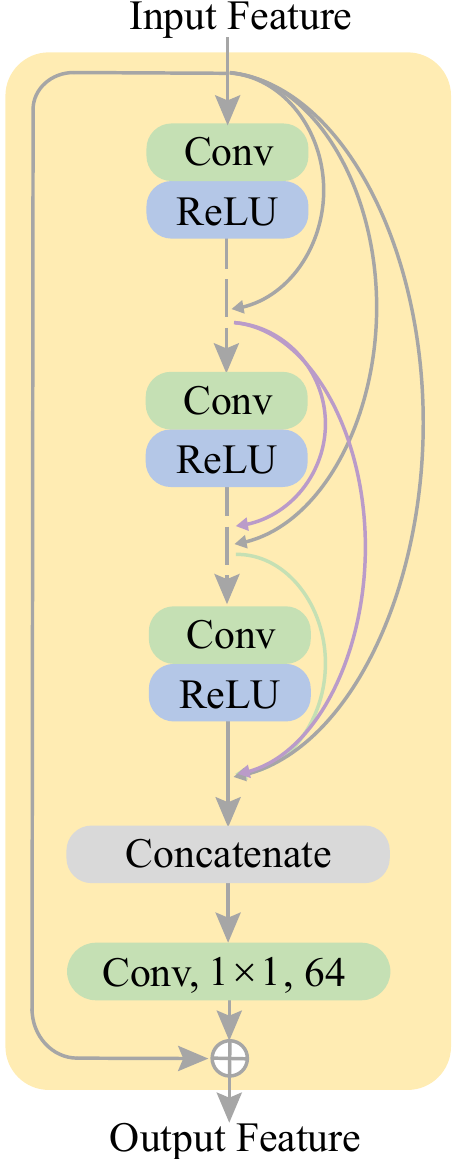}
		\\
		(a) Backbone Network  & 
		(b) Residual Dense Block \\
	\end{tabular}
	\vspace{-5pt}
	\caption{%
		\textbf{Architecture of the backbone network.}
		We use a DownShuffle layer in the backbone network to distribute the motion information into multiple channels.
		We use residual dense blocks to learn hierarchical features.
	}
	\label{fig:basic_module_architecture}
 	\vspace{-10pt} 
\end{figure}

    \Paragraph{Backbone Network.}
        We use a variation of the residual dense network~\cite{zhang2018residual} as the backbone network.
        As shown in~\figref{fig:basic_module_architecture},
        the backbone module consists of one DownShuffle layer and one UpShuffle layer~\cite{shi2016real}, six convolutional layers, and six residual dense blocks~\cite{zhang2018residual}. 
        The residual dense block consists of four $3\times 3$ convolutional layers, one $1 \times 1$ convolutional layer, and four ReLU activation layers.
        All of the hierarchical features from the residual dense blocks are concatenated for successive network modules.
    
    \Paragraph{Loss Function.}
       	Our loss function consists of two terms including the pixel reconstruction and cycle-consistency loss:
       	\begin{equation}\label{eqn:losscombination}
       	\mathcal{L} = \mathcal{L}_p + \mathcal{L}_c.
       	\end{equation}
       	\textit{Pixel reconstruction loss} $\mathcal{L}_p$ measures the overall pixel difference between the ground-truth frames $\mathbf{G}_{n}^t$ and the reconstructed frames $\mathbf{\hat{I}}_{n}^{t}$:
        \begin{equation}\label{pixelloss}
            \mathcal{L}_p = \frac{1}{T} \sum_{t=1}^{T}
            \sum_{n=1}^{2M-1} \rho\left( \mathbf{\hat{I}}_{n}^{t} - \mathbf{G}_{n}^t \right),
        \end{equation}
        where $\rho({x}) = \sqrt{{x}^2+\epsilon^2}$ is the Charbonnier penalty function~\cite{charbonnier1994two}.
        $T$ represents the iterations executed on the recurrent module.
        We use \textit{cycle consistency loss} $\mathcal{L}_c$ to ensure the spatial consistency between temporary inputs $\mathbf{I}_{n}^{'t}$ and the re-generated frames $\mathbf{\hat{I}}_{n}^{t}$ in the pyramid architecture:
        \begin{equation}\label{cycleloss}
            \mathcal{L}_c
            = \frac{1}{T} \sum_{t=1}^{T} \sum_{n\in {\Omega}} \rho 
            \left( \mathbf{I}_{n}^{'t} - \mathbf{\hat{I}}_{n}^{t} \right),
        \end{equation}
        where ${\Omega}$ is the index of all cycle-paired frames.

     \Paragraph{Training Dataset.}
        We use the Adobe240 dataset~\cite{su2017deep} to train the proposed network.
        It consists of 120 videos at 240 fps with the resolution of $1280 \times 720$.
        We use 112 of the videos to construct the training set.
        The following {discrete} degradation model is used to generate the training data:
        \begin{equation}
        \mathbf{B}_{2i} = \frac{1}{2 \tau + 1} \sum\nolimits_{j=iK-\tau}^{j = iK+\tau} \mathbf{L}_{j},  \quad i = 0,1,\dots,N,
        \label{bluemodel2}
        \end{equation}
        where $\mathbf{L}_i$ is the $i$-th high-frame-rate latent image, $\mathbf{B}_{2i}$ is the $i$-th acquired low-frame-rate blurred frame, the parameter $K$ determines the frame rate of acquired frames, $\leftright{2\tau + 1}$ corresponds to the equivalent long exposure time, that restricts the degree of blur~\cite{brooks2019learning}.
        We down-sample the high-frame-rate sequences to generate ground-truth frames.
        The frame rate of the ground-truth sequence is two times that of the blurry sequence.
        We use~\eqnref{bluemodel2} with parameters $K=8$ and $\tau=5$ to generate the training data.
        The resolution of training images is $640 \times 352$.
        Considering the computational complexity, we choose the temporal length of $T=2$.
        We augment the training data by horizontal and vertical flipping, randomly cropping as well as reversing the temporal order of the training samples.

    \Paragraph{Training Strategy.}
        We utilize the AdaMax~\cite{kingma2014adam} optimizer with parameters $\beta_1=0.9$ and $\beta_2=0.999$.
        We use a batch size of 2, and the initial learning rate is $1e^{-3}$. 
        We train the model for 40 epochs, then reduce the learning rate by a factor of 0.2 and fine-tune the entire model for another 5 epochs. 
        We train the network on an RTX-2080 Ti GPU card.
        It takes about two days to converge.

\section{Experimental Results}
    In this section, we first introduce the evaluation datasets and then conduct ablation studies to analyze the contribution of each proposed component.
    Finally, we compare the proposed model with state-of-the-art algorithms.

    \subsection{Evaluation Datasets and Metrics}\label{sec:metric}
        We evaluate the proposed model on two video datasets and measure the motion smoothness of the synthesized video sequences for a comprehensive understanding.

    \Paragraph{Adobe240.}
        We use $8$ videos of the Adobe240 dataset~\cite{su2017deep} for evaluation.
        Each video comes at 240 fps with the resolution of $1280 \times 720$.

        \Paragraph{YouTube240.}
        We download 59 slow-motion videos from the YouTube website to construct our YouTube240 evaluation dataset.
        The videos are of the same resolution and frame rate with Adobe240.
        For both Adobe240 and YouTube240 datasets, we use~\eqnref{bluemodel2} with parameter $K=8$ and $\tau=5$ to generate the evaluation data.
        All of the frames are resized to $640 \times 352$.

    \begin{figure}[t]
 	\footnotesize

	\centering
	\renewcommand{\tabcolsep}{0.7pt} %
	\renewcommand{\arraystretch}{0.5} %
	\begin{tabular}{cc}

\includegraphics[width=0.46 \linewidth]{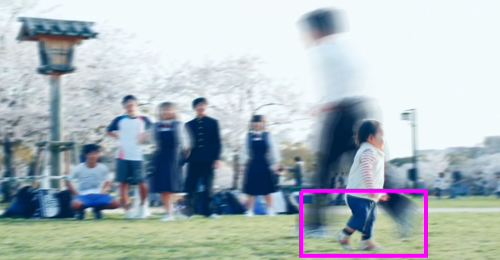} &
\includegraphics[width=0.46 \linewidth]{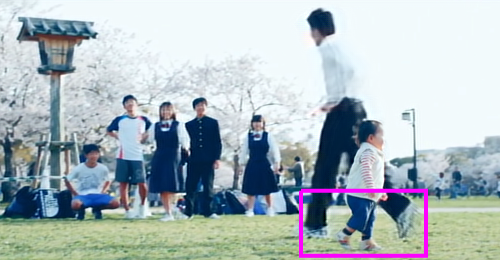}
\\
\includegraphics[width=0.46 \linewidth]{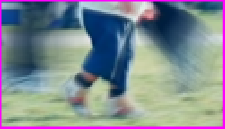} &
\includegraphics[width=0.46 \linewidth]{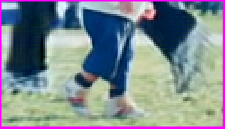}
\\

Inputs & \OursFirst 
\\
\includegraphics[width=0.46 \linewidth]{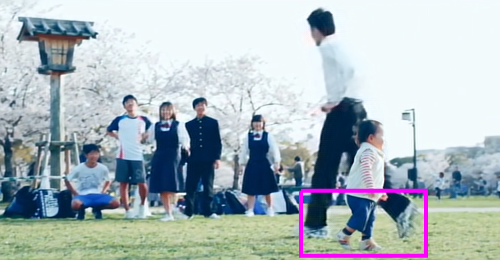} &
\includegraphics[width=0.46 \linewidth]{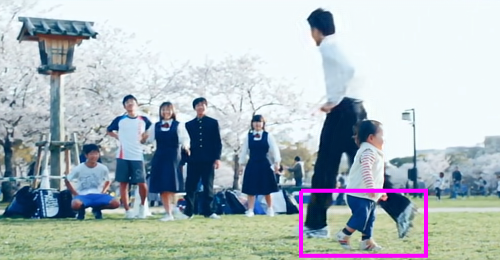}
\\
\includegraphics[width=0.46 \linewidth]{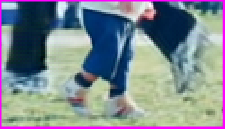} &
\includegraphics[width=0.46 \linewidth]{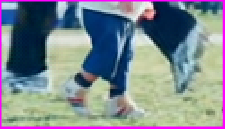}
\\
\OursSecond & {\OursThird} 

\end{tabular}
	\vspace{-5pt}
	\caption{
	 \textbf{Effect of network scales.}
    The model with larger scales can generate clear and sharper content.
	}
\vspace{-10pt}
\label{fig:avg_scale} %
\end{figure}

    \Paragraph{Motion Smoothness.}
        Our motion smoothness metric is based on optical flow estimation\cite{jin2019learning,sun2018pwc}.
        We first compute the differential optical flow $\mathbf{D}$ using three inputs $\mathbf{I}_{0:1:2}$ and three reference frames $\mathbf{R}_{0:1:2}$ with the following equation:
        \begin{equation} 
          \mathbf{D}  = 
          (\mathbf{F}_{\mathbf{I}_1\rightarrow \mathbf{I}_2} - \mathbf{F}_{\mathbf{I}_0 \rightarrow \mathbf{I}_1}) -
          (\mathbf{F}_{\mathbf{R}_1\rightarrow \mathbf{R}_2} - \mathbf{F}_{\mathbf{R}_0 \rightarrow \mathbf{R}_1}), 
        \end{equation}
        where $\mathbf{F}_{\mathbf{x} \rightarrow \mathbf{y}}$ is the estimated optical flow from frame $\mathbf{x}$ to frame $\mathbf{y}$. 
        We use the state-of-the-art PWC-Net~\cite{sun2018pwc} algorithm to estimate optical flow.
        PWC-Net integrates the classic pyramid processing, flow warping and cost volume filtering techniques into a convolutional neural network framework.
        Our motion smoothness metric is defined by: 
        \begin{equation}\label{eqn:smooth}
        	\mathcal{M}(s) =\log \sum_{d\in \mathbf{D}} \mathbf{1}_{[s,s+1)}\big(\big|\big|{d}\big|\big|_2\big) - \log |\mathbf{D}| 
        \end{equation}
        where $d$ denotes the 2-dimensional vector of matrix $\mathbf{D}$,
        $|\mathbf{x}|$ represents the size of the matrix $\mathbf{x}$,
        and the indicator function $\mathbf{1}_{\mathbf{A}}({x})$  equals to $1$ if $x$ belongs to set $\mathbf{A}$. 
        The $\mathcal{M}(s)$ measures the motion smoothness of three consecutive input frames concerning the pixel error length $s$, and lower $\mathcal{M}(s)$ indicates better performance.

    \subsection{Model Analysis}\label{ablation_study}
        To analyze the contributions of the proposed pyramid module,~\OursFullBModuleSmall,~ConvLSTM unit,~and cycle consistency loss, we perform the following extensive experiments:

     \begin{table}[t]
  \caption{
     \textbf{Analysis on network scales and recurrent module.}
     The numbers in \first{red} and \second{blue} represent the best and second-best results.
  }
  \vspace{-1mm}
  \label{tab:avg_scale}
  \footnotesize
  \renewcommand{\tabcolsep}{2.5pt} %
  \renewcommand{\arraystretch}{0.9} %
  \centering
  \begin{tabular}{ccccccc}
    \toprule
    
    \multirow{2}{*}[-0.28em]{Method}   
     & Runtime 
    & Parameters 
     &  \multicolumn{2}{c}{Adobe240~\cite{su2017deep}}
      & \multicolumn{2}{c}{YouTube240} \\
    
    \cmidrule(lr){4-5}
    \cmidrule(lr){6-7}
    & (seconds)  & (million) & PSNR & SSIM  &PSNR & SSIM \\
    \midrule
        \OursFirst \textit{ -w/o rec} %
              & 0.01 %
              & 2.27
              & 31.37 & 0.9129 
              & 34.10 & 0.9374\\ %
    
        \OursSecond \textit{ -w/o rec}
            & 0.06 %
            & 3.44
              & 31.67 & 0.9181
              & 34.54 & 0.9392 \\
       
        \OursThird \textit{ -w/o rec}
              & 0.12 %
              & 4.62
              & {32.06} & {0.9190} 
              & 34.72 & 0.9411\\ %
        \midrule
         \OursConvLstmFirst 
              & 0.02%
              & 2.29
              & 31.87 & 0.9183 
              & 34.41 & 0.9400\\ %
             
         \OursConvLstmSecond
            & 0.10 %
            & 3.49
              & \second{32.39} & \second{0.9212}
              & \second{34.77} & \second{0.9419} \\
              
        \OursThirdNewLoss 
            & 0.28
            & 4.68
            & \first{32.59} & \first{0.9258}
            & \first{35.10}& \first{0.9443} \\ 
        \bottomrule

  \end{tabular}
\end{table}

     \begin{table}[t]
  \caption{
     \textbf{Analysis on ConvLSTM unit.}
     We evaluate three variations, including the model using LSTM (\OursLstmFirst), the model using ConvLSTM (\OursConvLstmFirstY), and the model without using recurrent model (\OursFirstNa).
  }
  \label{tab:avg_conv}
  \footnotesize
  \renewcommand{\tabcolsep}{8pt} %
  \centering
  \begin{tabular}{lcccc}
    \toprule
    
    \multirow{2}{*}[-0.28em]{Method}   
     &  \multicolumn{2}{c}{Adobe240~\cite{su2017deep}}
      & \multicolumn{2}{c}{YouTube240} \\
    
    \cmidrule(lr){2-3}
    \cmidrule(lr){4-5}
    
    & PSNR & SSIM  &PSNR & SSIM \\
    \midrule
    
        \OursFirstNa
              & 30.39 & 0.8974
              & 33.32  & 0.9263 \\ %
       \OursLstmFirst
              & \second{31.38}  & \second{0.9120}
              & \second{34.07} & \second{0.9283} \\
        \OursConvLstmFirstY
              & \first{31.87} & \first{0.9183} 
              & \first{34.41}  & \first{0.9400}  \\
        \bottomrule

  \end{tabular}
\end{table}

    \begin{table*}[htb]
  \caption{
     \textbf{Quantitative comparisons on the Adobe240~\cite{su2017deep} and YouTube240 \textsc{evaluation} sets.} 
  }
  \label{tab:avg_compare_all}
  \footnotesize
  \renewcommand{\tabcolsep}{3.5pt} %
  \centering

  \begin{tabular}{lccccccccccccccc}
    \toprule

    \multirow{3}{*}[-0.58em]{Method} &  
    
    \multirow{1}{*}[-0.58em]{Runtime}  &
    
    \multirow{1}{*}[-0.58em]{Parameters}  &

     \multicolumn{4}{c}{Deblurring} &
     
     \multicolumn{4}{c}{Interpolation} & 
     
     \multicolumn{4}{c}{Comprehensiveness} \\
     
      \cmidrule(lr){4-7}
      \cmidrule(lr){8-11}
      \cmidrule(lr){12-15}
    &(seconds) & (million) & \multicolumn{2}{c}{Adobe240~\cite{su2017deep}} &
    \multicolumn{2}{c}{YouTube240}  &
    \multicolumn{2}{c}{Adobe240~\cite{su2017deep}} &
    \multicolumn{2}{c}{YouTube240}  
    &
    \multicolumn{2}{c}{Adobe240~\cite{su2017deep}} &
    \multicolumn{2}{c}{YouTube240}  
    \\
    
    \cmidrule(lr){4-5}
    \cmidrule(lr){6-7}
    \cmidrule(lr){8-9}
    \cmidrule(lr){10-11}
    \cmidrule(lr){12-13}
    \cmidrule(lr){14-15}
    &  &  &   PSNR & SSIM  &PSNR & SSIM 
    &   PSNR & SSIM  &PSNR & SSIM 
    &   PSNR & SSIM  &PSNR & SSIM \\

       \midrule
                
              Blurry Inputs
              & --- & ---
               & 28.68 & 0.8584 
              & 31.96 & 0.9119
              & --- & ---
              & --- & ---
              & --- & ---
              & --- & --- \\
       
              Super SloMo %
              & --- & 39.6
              & --- & ---
              & --- & ---
              & 27.52 & 0.8593 
              & 30.84 & 0.9107
              & --- & ---
              & --- & ---
              \\
        
              MEMC-Net %
              & ---  & 70.3
              & --- & ---
              & --- & ---
              & 30.83 & 0.9128
              & 34.91 & \second{0.9596} 
              & --- & ---
              & --- & ---\\

              DAIN
              & --- & 24.0 
              & --- & ---
              & --- & ---
              & \second{31.03} & \second{0.9172}
              &\second{35.06} & \first{0.9615}
              & --- & ---
              & --- & ---\\
        \midrule
              EDVR + Super SloMo %
              & 0.42 & 63.2
              & \multirow{3}{*}{\first{32.76}} %
              & \multirow{3}{*}{\first{0.9335}} %
              & \multirow{3}{*}{\second{34.66}}%
              & \multirow{3}{*}{\first{0.9448}}%
              & 27.79 & 0.8671
              & 31.15 & 0.9136
              & 30.28 & 0.9003
              & 32.91 & 0.9292 \\
              EDVR + MEMC-Net %
              & 0.27 & 93.9
              & %
              & %
              & %
              & %
              & 30.22 & 0.9058
              & 33.49 & 0.9367
              & 31.49 & 0.9197
              & 34.08 & 0.9408\\
              
              EDVR + DAIN %
              & 1.13 & 47.6
              & %
              & %
              & %
              & %
              & 30.28 & 0.9070
              & 33.53 & 0.9378
              & \second{31.52} & \second{0.9203}
              & \second{34.10} & \second{0.9413}
              \\ 
        \midrule
              
              SRN + Super SloMo  %
              & 0.27 & 47.7
              & \multirow{3}{*}{29.42} %
              & \multirow{3}{*}{0.8753}%
              & \multirow{3}{*}{32.00} %
              & \multirow{3}{*}{0.9118} %
              & 27.22 & 0.8454 
              & 30.42 & 0.8970
              & 28.32 & 0.8604
              & 31.21 & 0.9044\\
           
              SRN + MEMC-Net %
              & 0.22 & 78.4
              & %
              & %
              & %
              & %
              & 28.25 & 0.8625 
              & 31.60 & 0.9107
              & 28.84 & 0.8689 
              & 31.80 & 0.9113 \\

              SRN + DAIN   %
              & 0.79 & 32.1
              & %
              & %
              & %
              & %
              & 27.83 & 0.8562 
              & 31.15 & 0.9059
              & 28.63 & 0.8658
              & 31.58 & 0.9089 \\
        \midrule
              Jin~\cite{jin2019learning}%
              & 0.25 & 10.8 
              & 29.40 & 0.8734
            & 32.06 & 0.9119
              & 29.24 & 0.8754
              & 32.24 & 0.9140
              & 29.32 & 0.8744
              & 32.15 & 0.9130 \\
        \midrule
            \OursThirdNewLoss~(Ours)
            & 0.28  & 4.68
             & \second{32.67} & \second{0.9236} 
             & \first{35.10} & \second{0.9417} 
            & \first{32.51} & \first{0.9280}
            & \first{35.10}& 0.9468 
            & \first{32.59} & \first{0.9258}
            & \first{35.10} & \first{0.9443}\\ 
        \bottomrule

  \end{tabular}

\end{table*}

    \begin{figure*}[!]
	\scriptsize        
	\centering
	\renewcommand{\tabcolsep}{1.0pt} %
	\begin{tabular}{cccccccccc}

\multirow{2}{*}[5.30em]{\includegraphics[width=0.177\linewidth]{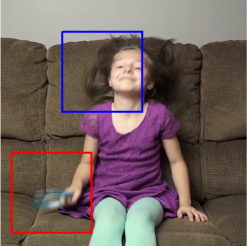}}&
\includegraphics[width=0.086\linewidth]{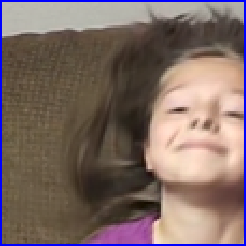}&
\includegraphics[width=0.086\linewidth]{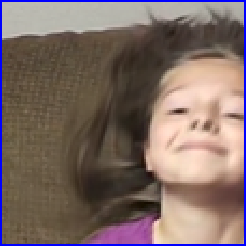}&
\includegraphics[width=0.086\linewidth]{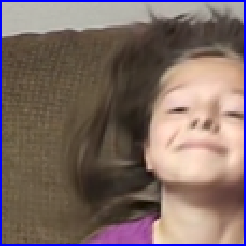}&
\includegraphics[width=0.086\linewidth]{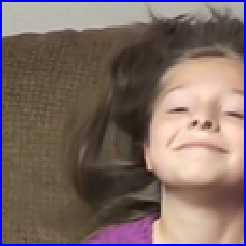}&
\includegraphics[width=0.086\linewidth]{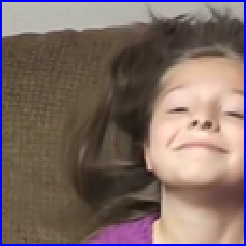}&
\includegraphics[width=0.086\linewidth]{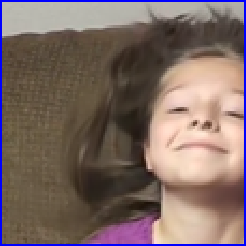}&
\includegraphics[width=0.086\linewidth]{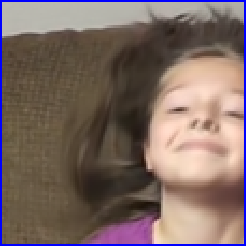}&
\includegraphics[width=0.086\linewidth]{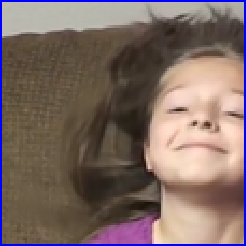}&
\includegraphics[width=0.086\linewidth]{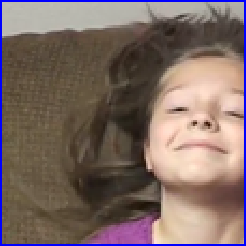}
\\
&
\includegraphics[width=0.086\linewidth]{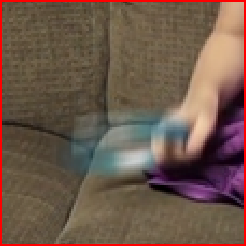}&
\includegraphics[width=0.086\linewidth]{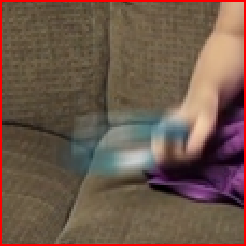}&
\includegraphics[width=0.086\linewidth]{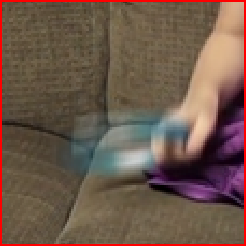}&
\includegraphics[width=0.086\linewidth]{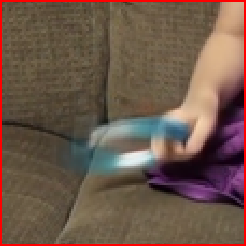}&
\includegraphics[width=0.086\linewidth]{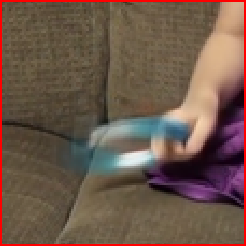}&
\includegraphics[width=0.086\linewidth]{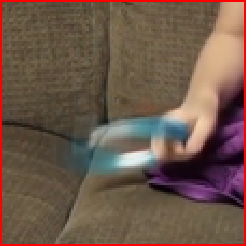}&
\includegraphics[width=0.086\linewidth]{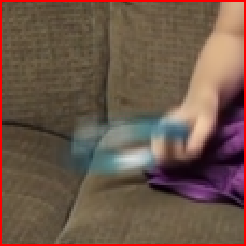}&
\includegraphics[width=0.086\linewidth]{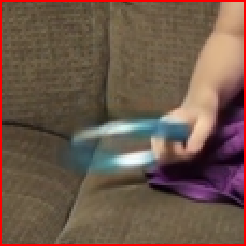}&
\includegraphics[width=0.086\linewidth]{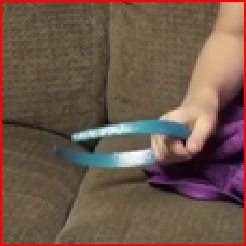}
\\

\multirow{2}{*}[5.30em]{\includegraphics[width=0.177\linewidth]{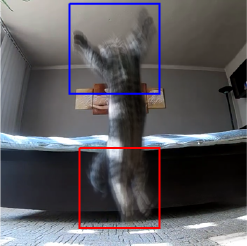}} &
\includegraphics[width=0.086\linewidth]{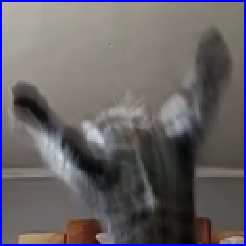}&
\includegraphics[width=0.086\linewidth]{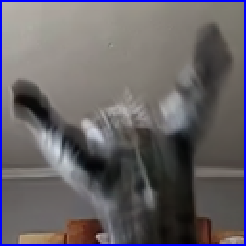}&
\includegraphics[width=0.086\linewidth]{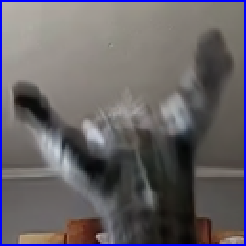}&
\includegraphics[width=0.086\linewidth]{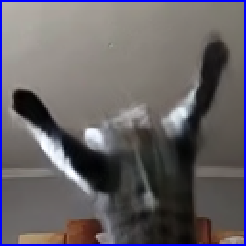}&
\includegraphics[width=0.086\linewidth]{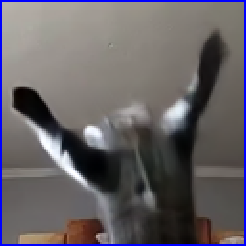}&
\includegraphics[width=0.086\linewidth]{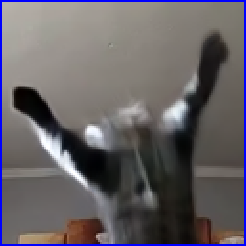}&
\includegraphics[width=0.086\linewidth]{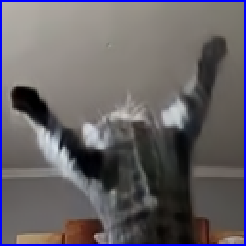}&
\includegraphics[width=0.086\linewidth]{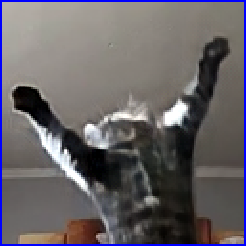}&
\includegraphics[width=0.086\linewidth]{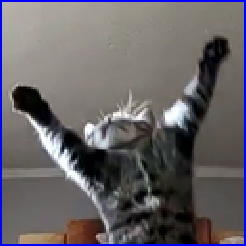}
\\
&
\includegraphics[width=0.086\linewidth]{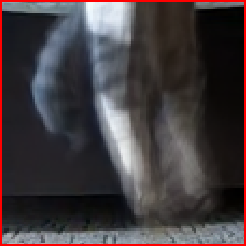}&
\includegraphics[width=0.086\linewidth]{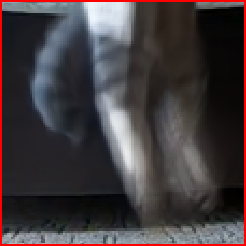}&
\includegraphics[width=0.086\linewidth]{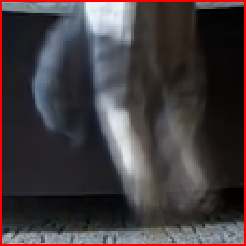}&
\includegraphics[width=0.086\linewidth]{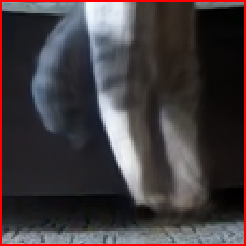}&
\includegraphics[width=0.086\linewidth]{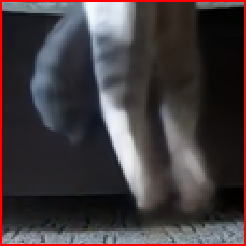}&
\includegraphics[width=0.086\linewidth]{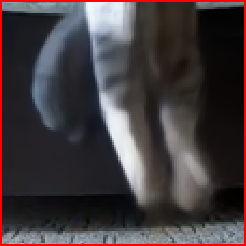}&
\includegraphics[width=0.086\linewidth]{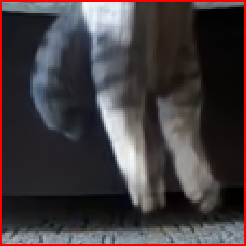}&
\includegraphics[width=0.086\linewidth]{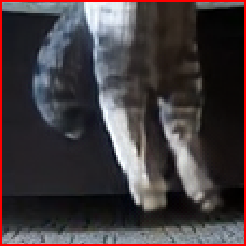}&
\includegraphics[width=0.086\linewidth]{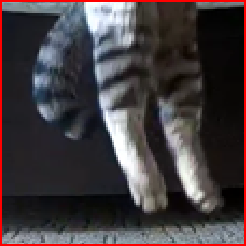}
\\
Blurry Inputs&
SRN\! + \!S.S. &
SRN\! + \!M.N. &
SRN\! + \!DAIN &
EDVR\! + \!S.S. &
EDVR\! + \!M.N. &
EDVR\! + \!DAIN&
Jin~\cite{jin2019learning}&
\OursThirdNewLoss(Ours) &
GT
\end{tabular}
	\caption{
       \textbf{Visual comparisons on the YouTube240 \textsc{evaluation} set.}
	   The pictures in the first two rows and the last two rows show the deblurred frames and the interpolated frames, respectively.
	   Our method generates clearer and sharper content.
       S.S. is short for Super SloMo~\cite{jiang2018super} and M.N. is short for MEMC-Net~\cite{bao2018memc}.
	}
	\label{fig:compare} %
	\vspace{-5pt}
\end{figure*}

   \Paragraph{Architecture Scalability.} 
        We first investigate the scalability of the pyramid module by evaluating networks with three different scales (\OursFirst,~\OursSecond,~\OursThird).
        We show the quantitative results in~\tabref{tab:avg_scale}, and provide the visual comparisons in~\figref{fig:avg_scale}.
        We find that the module using larger scales generates more clear details in~\figref{fig:avg_scale}.
        We observe that with the parameters of \OursHalfS~increasing from 2.29, 3.49 to 4.68 million, the networks steadily obtain better PSNR results from 31.87dB, 32.39dB to 32.59dB on the Adobe240 dataset.
        However, the runtime costs also increase from 0.02, 0.10, to 0.28 seconds.
        The comparisons show that the pyramid module is scalable, where the scales balance the computational complexity (execution time and model parameters) and restoration quality.

     \Paragraph{Inter-Pyramid Recurrent Module.}
        We then study the contributions of the proposed recurrent module by evaluating the model using recurrent modules and the model without using recurrent modules~(i.e., $\text{BIN}_{{l}}$ versus $\text{BIN}_{{l}}$~\textit{-w/o rec}, $l=2,3,4$).
        In~\tabref{tab:avg_scale}, we find that \OursThirdNewLoss~obtains a better SSIM of 0.9258 than the SSIM of 0.9212 achieved by~\OursThird~\textit{-w/o rec} in the Adobe240 set.
        The model using the recurrent module improves the restoration performance, it achieves about 0.5dB gain in the Adobe240 set and 0.3dB gain in the YouTube240 set.

    \Paragraph{ConvLSTM Module.}
        To analyze the contribution of ConvLSTM unit, we evaluate the model using LSTM (\OursLstmFirst), ConvLSTM (\OursConvLstmFirstY), and that without using any recurrent unit (\OursFirstNa).
        The \OursFirstNa~directly concatenates previous frames to propagate information recurrently.
        The results in~\tabref{tab:avg_conv}~show that the ConvLSTM unit performs better than the LSTM unit as well as the model without using recurrent unit.
        The ConvLSTM unit provides about 0.49dB PSNR gain in the Adobe240 set and 0.34dB gain in the YouTube240 set.

\begin{figure}[t]
	\footnotesize

	\centering
	\includegraphics[width= 0.475\textwidth]{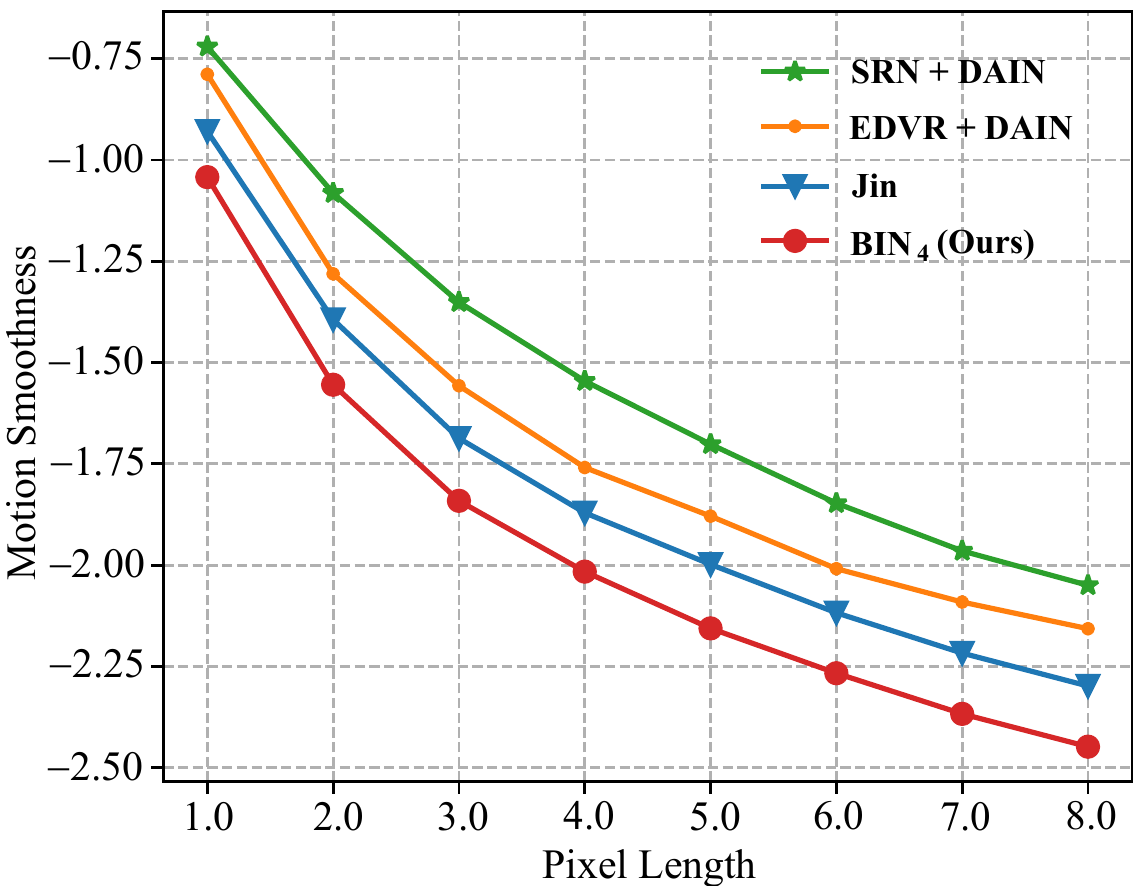}
	
	\caption{
		\textbf{Motion smoothness comparisons on the Adobe240 \textsc{evaluation} set.}
		The proposed model reaches the best performance in terms of motion smoothness.
        The smoothness metric is displayed in logarithm to show the difference.
	}
	\label{fig:smooth_ablation} 
\end{figure}

     \Paragraph{Cycle Consistency Loss.}
        Finally, we compare the model with cycle loss (\OursConvLstmThirdwithCyc) versus the model without cycle loss (\OursConvLstmThird).
        On the Adobe240 dataset, the PSNR of the model w/ and w/o cycle loss is 32.59dB and 32.42dB, respectively. Namely, the cycle loss provides $0.17$dB gain.
        The results demonstrate that the cycle loss ensures consistency of frames and it helps the model to generate fine details of moving objects.

    \subsection{Compare with the State-of-the-arts}
        We evaluate the proposed method against the algorithm proposed by~Jin~\textit{\etal}~\cite{jin2019learning}. 
        Their model synthesizes nine intermediate frames using two blurred inputs.
        We extract the center interpolated frame to compare with our results.
        Besides, we construct several cascade methods by connecting deblurring and interpolation models,~including
        EDVR~\cite{wang2019edvr}, SRN~\cite{tao2018scale} for deblurring,~and Super~SloMo~\cite{jiang2018super}, MEMC~\cite{bao2018memc}, DAIN~\cite{bao2019depth} for interpolation.
        We compare our model with the state-of-the-art algorithms in the following aspects:

    \Paragraph{Interpolation Evaluation.}
        As shown in~\tabref{tab:avg_compare_all} and~\figref{fig:compare} , our model~performs favorably against all the compared methods.
        Moreover, we find that our model performs better than the frame interpolation method using \textit{sharp} frames (\eg DAIN).
        For example, the PSNR of our model is 32.51dB, while the PSNR of DAIN is 31.03dB on the Adobe240 dataset.
        The main reason is that one blurred frame contains information of multiple sharp frames, and our method synthesizes the intermediate frame using several blurred frames, while the interpolation method only uses two sharp frames.
        Therefore, our model can exploit more space-time information from multiple blurred frames, resulting in more satisfactory intermediate frames.
    \Paragraph{Deblurring Evaluation.}
        We then compare the deblurring aspects with the state-of-the-art methods.
        As shown in~\tabref{tab:avg_compare_all}, our model performs slightly inferior to the state-of-the-art EDVR algorithm. 
        Our model achieves 0.09dB less than EDVR in terms of PSNR, but our model size (4.68 million) is much smaller than that of the EDVR (23.6 million), and our model requires less execution time.
     \Paragraph{Comprehensive Evaluation.}
        We compare the comprehensive performance of deblurring and interpolation.
        A high-performance pre-deblurring model in the cascade method helps the subsequent interpolation network to restore better results.
        As shown in~\tabref{tab:avg_compare_all}, the SRN model performs slightly inferior to the EDVR. 
        Thus the EDVR + DAIN has a better performance than SRN + DAIN.
        However, the best-performing cascade method (EDVR + DAIN) is still sub-optimal in terms of the overall performance.
        The overall PSNR of EDVR + DAIN is 31.52dB, while our model obtains PSNR of 32.59dB on the Adobe240 dataset.

        Compared to Jin~\textit{\etal}~\cite{jin2019learning}'s method, our approach obtains up to 3.27dB gain on the Adobe240 dataset.
        Their training dataset has less fast-moving screens and camera shakes than the Adobe240 dataset.
        Therefore, the Adobe240 dataset has a severer blur than Jin's training dataset.
        We note that Jin~\textit{\etal} do not publish their training code at the time of submission.
        We cannot optimize their model on the Adobe240 dataset for fair comparisons.
        Nevertheless, compared with their method, our network benefits from scalable structure and recurrent information propagation, thus obtains significant performance gains.

      \begin{figure}[t]
 	\footnotesize

	\centering
	\renewcommand{\tabcolsep}{0.7pt} %
	\begin{tabular}{ccc}

\parbox[t]{2.1mm}{\multirow{1}{*}[5.6em]{\rotatebox[origin=c]{90}{SRN + DAIN}}} &
\includegraphics[width=0.45 \linewidth]{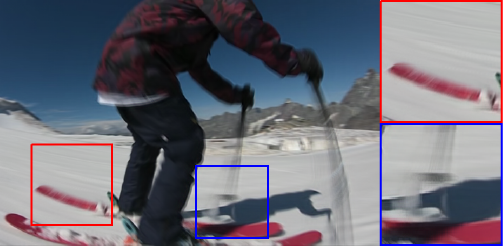} &
\includegraphics[width=0.45 \linewidth]{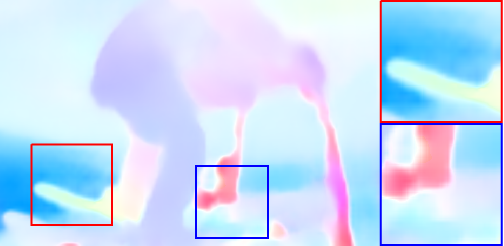}
\\
\parbox[t]{2.1mm}{\multirow{1}{*}[6.0em]{\rotatebox[origin=c]{90}{EDVR + DAIN}}} &
\includegraphics[width=0.45 \linewidth]{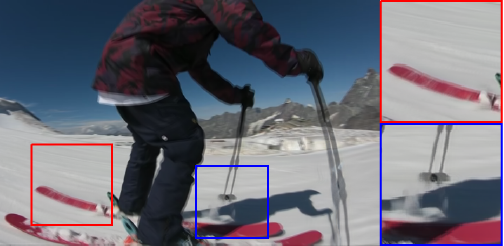} &
\includegraphics[width=0.45 \linewidth]{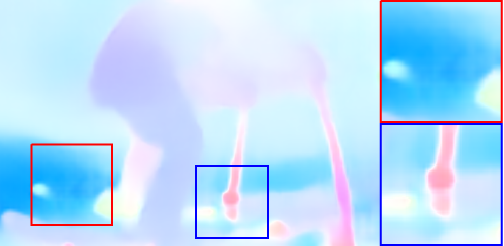}
\\
\parbox[t]{2.1mm}{\multirow{1}{*}[4.5em]{\rotatebox[origin=c]{90}{Jin~\cite{jin2019learning}}}} &
\includegraphics[width=0.45 \linewidth]{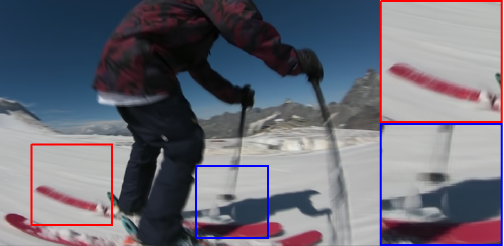} &
\includegraphics[width=0.45 \linewidth]{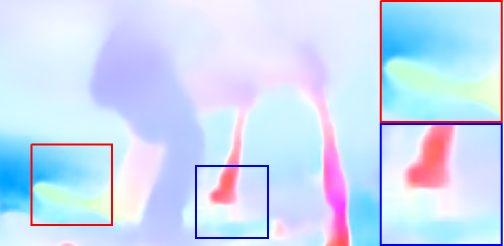}
\\
\parbox[t]{2.1mm}{\multirow{1}{*}[5.5em]{\rotatebox[origin=c]{90}{\OursThirdNewLoss(Ours)}}} &
\includegraphics[width=0.45 \linewidth]{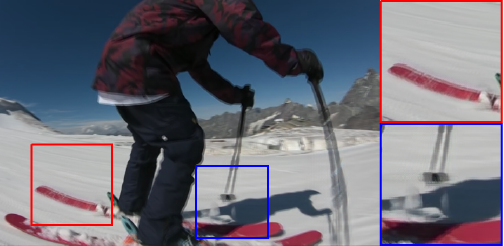} &
\includegraphics[width=0.45 \linewidth]{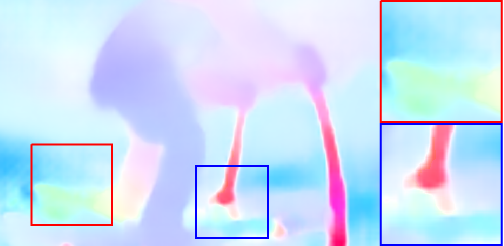}
\\
& 
Overlapped frames & Optical flow

\end{tabular}
	\caption{
	 \textbf{Visual comparisons on the YouTube240 dataset.}
    We use the PWC-net~\cite{sun2018pwc} to estimate optical flow of two adjacent output frames.
    The optical flow of our model has smoother shapes.
	}
\vspace{-10pt}
\label{fig:flow_map} %
\end{figure}

    \Paragraph{Motion Smoothness Evaluation.}
        We compare the motion smoothness performance based on the metric introduced in~\secref{sec:metric}.
        A lower metric indicates a better performance.
        As shown in~\figref{fig:smooth_ablation},  Jin~\cite{jin2019learning}'s model performs favorably against all the cascade methods (we show SRN + DAIN and EDVR + DAIN for brevity), and our algorithm has a better smoothness metric than Jin's model.
        In~\figref{fig:flow_map}, the optical flow of our model has smoother shapes compared with the cascade methods.
        Our network is a unified model with a broad temporal scope, which helps generate smooth frames.
        Besides, compared to the approximate recurrent mechanism of~Jin~\cite{jin2019learning}, our proposed~\OursFullBModuleSmall~adopts ConvLSTM cells to propagate the frame information across time.
        It can further enforce temporal consistency between the deblurred and interpolated frames. 
        Thus, our method is superior to all the cascade methods and Jin's model.

\section{Conclusion}
    In this work, we propose a blurry video frame interpolation method to address the joint video enhancement problem. 
    Our model consists of pyramid modules and inter-pyramid recurrent modules.
    The pyramid module is scalable, where the scales balance the computational complexity and restoration quality.
    We use the cycle consistency loss to ensure the consistency of inter-frames in the pyramid module.
    Furthermore, the~\OursFullBModuleSmall~utilizes the spatial-temporal information to generate temporally smoother results.
    Extensive quantitative and qualitative evaluations demonstrate that the proposed method performs favorably against the existing approaches.

\small{
	\Paragraph{Acknowledgment}. 
	To be add.
	}

\newpage
{
\small
\bibliographystyle{ieee}
\bibliography{mybib}
}

\end{document}